\documentclass{article}
\usepackage{ijcai21}

\usepackage{times}
\usepackage{soul}
\usepackage{url}
\usepackage[hidelinks]{hyperref}
\usepackage[utf8]{inputenc}
\usepackage[small]{caption}
\usepackage{graphicx}
\usepackage{amsmath}
\usepackage{amsthm}
\usepackage{booktabs}
\usepackage{algorithm}  
\usepackage{amsfonts}
\usepackage{algpseudocode} 
\newtheorem{assumption}{Assumption}
\newtheorem{corollary}{Corollary}

\newtheorem{theorem}{Theorem}

\title{Towards Heterogeneous Clients with Elastic Federated Learning}

\author{
 Zichen Ma$^{1,2}$
 \and
 Yu Lu$^{1,2}$\and
 Zihan Lu$^{3}$\and
 Wenye Li$^{1}$\And
 Jinfeng Yi$^2$\and
 Shuguang Cui$^1$
\affiliations
 $^1$The Chinese University of Hong Kong, Shenzhen\\
 $^2$JD AI Lab\\
 $^3$Ping An Technology\\
}

\begin{document}

\maketitle

\begin{abstract}
Federated learning involves training machine learning models over devices or data silos, such as edge processors or data warehouses, while keeping the data local. Training in heterogeneous and potentially massive networks introduces bias into the system, which is originated from the non-IID data and the low participation rate in reality. In this paper, we propose Elastic Federated Learning (EFL), an unbiased algorithm to tackle the heterogeneity in the system, which makes the most informative parameters less volatile during training, and utilizes the incomplete local updates. It is an efficient and effective algorithm that compresses both upstream and downstream communications. Theoretically, the algorithm has convergence guarantee when training on the non-IID data at the low participation rate. Empirical experiments corroborate the competitive performance of EFL framework on the robustness and the efficiency.
\end{abstract}

\section{Introduction}
\label{Section1}
Federated learning (FL) has been an attractive distributed machine learning paradigm where participants jointly learn a global model without data sharing \cite{mcmahan2017communication}. It embodies the principles of focused collection and data minimization, and can mitigate many of the systemic privacy risks and costs resulting from traditional, centralized machine learning \cite{kairouz2019advances}. While there are plenty of works on federated optimization, bias in the system still remains a key challenge. The origins of bias are from (\romannumeral1) the statistical heterogeneity that data are not independent and identically distributed (IID) across clients (\romannumeral2) the low participation rate that is due to limited computing and communication resources, e.g., network condition, battery, processors, etc. 

Existing FL methods empower participants accomplish several local updates, and the server will abandon struggle clients, which attempt to alleviate communication burden. The popular algorithm, FedAvg \cite{mcmahan2017communication}, first allows clients to perform a small number of epochs of local stochastic gradient descent (SGD), then successfully completed clients communicate their model updates back to the server, and stragglers will be abandoned.

While there are many variants of FedAvg and they have shown empirical success in the non-IID settings, these algorithms do not fully address bias in the system. The solutions are sub-optimal as they either employ a small shared global subset of data \cite{zhao2018federated} or greater number of models with increased communication costs \cite{karimireddy2020scaffold,li2018federated,li2019convergence}. Moreover, to the best of our knowledge, previous models do not consider the low participation rate, which may restrict the potential availability of training datasets, and weaken the applicability of the system.

In this paper, we develop Elastic Federated Learning (EFL), which is an unbiased algorithm that aims to tackle the statistical heterogeneity and the low participation rate challenge. 

\textbf{Contributions} of the paper are as follows:
Firstly, EFL is robust to the non-IID data setting. It incorporates an elastic term into the local objective to improve the stability of the algorithm, and makes the most informative parameters, which are identified by the Fisher information matrix, less volatile. Theoretically, we provide the convergence guarantees for the algorithm. 

Secondly, even when the system is in the low participation rate, i.e., many clients may be inactive or return incomplete updates, EFL still converges. It utilizes the partial information by scaling the corresponding aggregation coefficient. We show that the low participation rate will not impact the convergence, but the tolerance to it diminishes as the learning continuing.

Thirdly, the proposed EFL is a communication-efficient algorithm that compresses both upstream and downstream communications. We provide the convergence analysis of the compressed algorithm as well as extensive empirical results on different datasets. The algorithm requires both fewer gradient evaluations and communicated bits to converge.
\section{Related Work}
\textbf{Federated Optimization} 
Recently we have witnessed significant progress in developing novel methods that address different challenges in FL; see \cite{kairouz2019advances,li2020federated}. In particular, there have been several works on various aspects of FL, including preserving the privacy of users \cite{duchi2014privacy,mcmahan2017learning,agarwal2018cpsgd,zhu2020federated} and lowering communication cost \cite{reisizadeh2020fedpaq,dai2019hyper,basu2019qsparse,li2020acceleration}. Several works develop algorithms for the homogeneous setting, where the data samples of all users are sampled from the same probability distribution \cite{stich2018local,wang2018cooperative,zhou2017convergence,lin2018don}. More related to our paper, there are several works that study statistical heterogeneity of users’ data samples in FL \cite{zhao2018federated,sahu2018convergence,karimireddy2020scaffold,haddadpour2019convergence,li2019convergence,khaled2020tighter}, but the solutions are not optimal as they either violate privacy requirements or increase the communication burden.

\textbf{Lifelong Learning} The problem is defined as learning separate tasks sequentially using a single model without forgetting the previously learned tasks. In this context, several popular approaches have been proposed such as data distillation \cite{parisi2018lifelong}, model expansion \cite{rusu2016progressive,draelos2017neurogenesis}, and memory consolidation \cite{soltoggio2015short,shin2017continual}, a particularly successful one is EWC \cite{kirkpatrick2017overcoming}, a method to aid the sequential learning of tasks.

To draw an analogy between federated learning and the problem of lifelong learning, we consider the problem of learning a model on each client in the non-IID setting as a separate learning problem. In this sense, it is natural to use similar tools to alleviate the bias challenge. While two paradigms share a common main challenge in some context, learning tasks in lifelong learning are serially carried rather than in parallel, and each task is seen only once in it, whereas there is no such limitation in federated learning.

\textbf{Communication-efficient Distributed Learning} A wide variety of methods have been proposed to reduce the amount of communication in distributed machine learning. The substantial existing research focuses on (\romannumeral1) communication delay that reduces the communication frequency by performing local optimization \cite{konevcny2016federated,mcmahan2017communication} (\romannumeral2) sparsification that reduces the entropy of updates by restricting changes to only a small subset of parameters \cite{aji2017sparse,tsuzuku2018variance} (\romannumeral3) dense quantization that reduces the entropy of the weight updates by restricting all updates to a reduced set of values \cite{alistarh2017qsgd,bernstein2018signsgd}.

Out of all the above-listed methods, only \emph{FedAvg} and \emph{signSGD} compress both upstream and downstream communications. All other methods are of limited utility in FL setting, as they leave communications from the server to clients uncompressed.

\section{Elastic Federated Learning}

\begin{algorithm}[tb]
  \caption{EFL. $N$ clients are indexed by $k$; $B$ is the local mini-batch size, $p^k$ is the probability of the client is selected ,$\eta_{\tau}$ is the learning rate, $E$ is the maximum number of time steps each round has, and $0\leq s_{\tau}^k\leq E$ is the number of local updates the client completes in the $\tau$-th round.}  
  \label{AL1}  
  \textbf{Server executes:}
  \begin{algorithmic}  
    \State initialize $\omega_0$, each client $k$ is selected with probability $p^k$, $R_0^G, R_0^k\gets$ 0.
    \For{each round $\tau$ = 1,2, ...}
      \State $S_{\tau}$ $\gets$ (random subset of $N$ clients)
      \For{each client $k$ $\in S_t$ \textbf{in parallel}}
      \State $\Delta\omega_{\tau E}^k, u_{\tau}^k, v_{\tau}^k\gets$ ClientUpdate($k$, $ST(\Delta\omega_{(\tau-1) E}^G)$, $\sum_k u_{\tau-1}^k,\sum_k v_{\tau-1}^k$)
      \State $\Delta\omega_{\tau E}^G=R_{(\tau-1)E}^G+\sum_kp_{\tau}^k\Delta\omega_{\tau E}^k$
      \State $R_{\tau E}^G=\Delta\omega_{\tau E}^G-ST(\Delta\omega_{\tau E}^G)$
      \State $\omega_{(\tau+1) E}^G = \omega_{\tau E}^G+\Delta\omega_{\tau E}^G$
      \EndFor
      \State return $ST(\Delta\omega_{\tau E}^G),\sum_k u_{\tau}^k,\sum_k v_{\tau}^k$ to participants
     \EndFor  
  \end{algorithmic}  
  \textbf{ClientUpdate($k$, $ST(\Delta\omega_{(\tau-1) E}^G)$, $\sum_k u_{\tau-1}^k,\sum_k v_{\tau-1}^k$):}
  \begin{algorithmic} 
    \State $\xi$ $\gets$ split local data into batches of size $B$
    \For{batch $\xi$ $\in B$}
    \State $\omega_{\tau E}^k=\omega_{(\tau-1) E}^k+\Delta\omega_{(\tau-1) E}^G$
    \For{$j=0,...,s_{\tau}^k-1$}
    \State $\omega_{\tau E+j+1}^k=\omega_{\tau E+j}^k-\eta_{\tau}g_{\tau E+j}^k$
    \EndFor
    \State $\Delta\omega_{\tau E}^k=R_{(\tau-1) E}^k+\omega_{\tau E+s_{\tau}^k}^k-\omega_{\tau E}^k$
    \State $R_{\tau E}^k=\Delta\omega_{\tau E}^k-ST(\Delta\omega_{\tau E}^k)$
    \State $u_{\tau}^k=diag(I_{\tau,k})$
    \State $v_{\tau}^k=diag(I_{\tau,k})\omega_{\tau E+s_{\tau}^k}^k$
    \EndFor
    \State return $\Delta\omega_{\tau E}^k, u_{\tau}^k, v_{\tau}^k$ to the server
  \end{algorithmic}  
\end{algorithm}

\subsection{Problem Formulation}
EFL is designed to mitigate the heterogeneity in the system, where the problem is originated from the non-IID data across clients and the low participation rate. In particular, the aim is to minimize:
\begin{equation}
    \label{E1}
    \min \limits_{\omega} F(\omega)=\sum_{k=1}^N p^k\widetilde{F}_k(\omega),
\end{equation}
where $N$ is the number of participants, $p^k$ denotes the probability of $k$-th client is selected. 
Here $\omega$ represents the parameters of the model, and $\widetilde{F}_k(\omega)$ is the local objective of $k$-th client. 

Assuming there are at most $T$ rounds. For the $\tau$-th round, the clients are connected via a central aggregating server, and seek to optimize the following objective locally:
\begin{equation}
    \label{E2}
    \widetilde{F}_{\tau,k}(\omega)=f_k(\omega)+\frac{\lambda}{2}\sum_{i=1}^N(\omega-\omega_{\tau-1}^i)^Tdiag(I_{\tau-1,i})(\omega-\omega_{\tau-1}^i),
\end{equation}
where $f_k(\omega)$ is the local empirical risk over all available samples at $k$-th client. $\omega_{\tau-1}^i$ is the model parameters of $i$-th client in the $(\tau-1)$-th round. $I_{\tau-1,i}=I(\omega_{\tau-1}^i)$ is the Fisher information matrix, which is the negative expected Hessian of log likelihood function, and $diag(I_{\tau-1,i})$ is the matrix that preserves values of diagonal of the Fisher information matrix, which aims to penalize parts of the parameters that are too volatile in a round.

We propose to add the elastic term (the second term of Equation (\ref{E2})) to the local subproblem to restrict the most informative parameters' change. It alleviates bias that is originated from the non-IID data and stabilizes the training. Equation (\ref{E2}) can be further rearranged as
\begin{equation}
    \label{E3}
    \begin{split}
    \widetilde{F}_{\tau,k}(\omega)&=f_k(\omega)+\frac{\lambda}{2}\omega^T\sum_{i=1}^Ndiag(I_{\tau-1,i})\omega\\
    &-\lambda\omega^T\sum_{i=1}^Ndiag(I_{\tau-1,i})\omega_{\tau-1}^i+Z,
    \end{split}
\end{equation}
where $Z$ is a constant. Let $u_{\tau-1}^k=diag(I_{\tau-1,k})$, and $v_{\tau-1}^k=diag(I_{\tau-1,k})\omega_{\tau-1}^k$.

Suppose $\omega^*$ is the minimizer of the global objective $F$, and denote by $\widetilde{F}_k^*$ the optimal value of $\widetilde{F}_k$. We further define the degree to which data at $k$-th client is distributed differently than that at other clients as $D_k=\widetilde{F}_k(\omega^*)-\widetilde{F}_k^*$, and $D=\sum_{k=1}^Np^kD_k$. We consider discrete time steps $t=0, 1, ...$. Model weights are aggregated and synchronized when $t$ is a multiple of $E$, i.e., each round consists of $E$ time steps. In the $\tau$-th round, EFL, presented in Algorithm \ref{AL1}, executes the following steps:

Firstly, the server broadcasts the compressed latest global weight updates $ST(\Delta\omega_{(\tau-1) E}^G)$, $\sum_k u_{\tau-1}^k$, and $\sum_k v_{\tau-1}^k$ to participants. Each client then updates its local weight: $\omega_{\tau E}^k=\omega_{(\tau-1) E}^k+\Delta\omega_{(\tau-1) E}^G$.

Secondly, each client runs SGD on its local objective $\widetilde{F}_k$ for $j=0,...,s_{\tau}^k-1$:
\begin{equation}
    \label{E4}
    \omega_{\tau E+j+1}^k=\omega_{\tau E+j}^k-\eta_{\tau}g_{\tau E+j}^k,
\end{equation}
where $\eta_{\tau}$ is a learning rate that decays with $\tau$, $0\leq s_{\tau}^k\leq E$ is the number of local updates the client completes in the $\tau$-th round, $g_t^k=\nabla \widetilde{F}_k(\omega_t^k,\xi_t^k)$ is the stochastic gradient of $k$-th client, and $\xi_t^k$ is a mini-batch sampled from client $k$'s local data. $\overline{g}_t^k=\nabla \widetilde{F}_k(\omega_t^k)$ is the full batch gradient at client $k$, and $\overline{g}_t^k=\mathbb{E}_{\xi_t^k}[g_t^k]$, $\Delta\omega_{\tau E}^k=R_{(\tau-1) E}^k+\omega_{\tau E+s_{\tau}^k}^k-\omega_{\tau E}^k$, where each client computes the residual as
\begin{equation}
    \label{E5}
    R_{\tau E}^k=\Delta\omega_{\tau E}^k-ST(\Delta\omega_{\tau E}^k).
\end{equation}
$ST(\cdot)$ is the compression method presented in Algorithm \ref{AL2}. Client sends the compressed local updates $ST(\Delta\omega_{\tau E}^k)$, $u_{\tau}^k$, and $v_{\tau}^k$ back to the coordinator.

Thirdly, the server aggregates the next global weight as
\begin{equation}
    \label{E6}
    \begin{split}
    \omega_{(\tau+1) E}^G &= \omega_{\tau E}^G+\Delta\omega_{\tau E}^G\\
    &=\omega_{\tau E}^G+R_{(\tau-1) E}^G+\sum_{k=1}^Np_{\tau}^k\Delta\omega_{\tau E}^k\\
    &=\omega_{\tau E}^G+R_{(\tau-1) E}^G-\sum_{k=1}^Np_{\tau}^k\sum_{j=0}^{s_{\tau}^k}\eta_{\tau}g_{\tau E+j}^k,
    \end{split}
\end{equation}
where $R_{\tau E}^G=\Delta\omega_{\tau E}^G-ST(\Delta\omega_{\tau E}^G)$. 

\begin{algorithm}[tb]
  \caption{Compression Method $ST$. $q$ is the sparsity, tensor $T \in \mathbb{R}^n$, $\hat{T}\in \{-\mu,0,\mu \}^n$}  
  \label{AL2}  
  \textbf{ST($T$):}
  \begin{algorithmic} 
    \State $k= \max(nq,1)$; $e= top_{k}(|T|)$
    \State $mask=(|T|\geq e)\in \{0,1\}^n$; $T^{mask}=mask\times T$
    \State $\mu=\frac{1}{k}\sum_{i=1}^n|T_i^{mask}|$
    \State $\hat{T}=\mu \times sign(T^{mask})$
    \State return $\hat{T}$
  \end{algorithmic}  
\end{algorithm}

As mentioned in Section \ref{Section1}, clients' low participation rate in a federated machine learning system is common in reality. EFL mainly focuses on two situations that lead to the low participation rate, which are not yet well discussed previously: (\romannumeral1) incomplete clients that can only submit partially complete updates (\romannumeral2) inactive clients that cannot respond to the server.

The client $k$ is inactive in the $\tau$-th round if $s_{\tau}^k=0$, i.e. it does not perform the local training, and the client $k$ is incomplete if $0<s_{\tau}^k<E$. $s_{\tau}^k$ is a random variable that can follow an arbitrary distribution. It can generally be time-varying, i.e., it may follow different distributions at different time steps. EFL also allows the aggregation coefficient $p_{\tau}^k$ to vary with $\tau$, and in the next subsection, we explore different schemes of choosing $p_{\tau}^k$ and their impacts on the model convergence.

\begin{table*}[tb]
\centering
\caption{Number of communication rounds required to reach $\epsilon$-accuracy. SC refers to strongly convex, NC refers to non-convex, and $\delta$ in MIME bounds Hessian dissimilarity. EFL preserves the optimal statistical rates (first term in SCAFFOLD) while improves the optimization.}
\begin{tabular}{cccc} 
\toprule
Algorithm & Bounded gradient & Convexity & \# Com. rounds \\
    \midrule
    SCAFFOLD & \checkmark & $\mu$-SC & $\frac{G^2}{\mu S\epsilon}+\frac{G}{\mu\sqrt{\epsilon}}+\frac{L}{\mu}$\\
    MIME  & \checkmark & $\mu$-SC & $\frac{G^2}{\mu S\epsilon}+\frac{\delta}{\mu}$\\
    VRL-SGD & $\times$ & NC & $\frac{N\sigma^2}{S\epsilon^2}+\frac{N}{\epsilon}$\\
    FedAMP & \checkmark & NC & $\frac{G^2}{L S\epsilon}+\frac{G}{\epsilon^{\frac{3}{2}}}+\frac{L^2}{\sqrt{\epsilon}}$\\
    EFL& \checkmark & $\mu$-SC & $\frac{G^2}{\mu S\epsilon}+\frac{L}{\mu\sqrt{\epsilon}}$\\
\bottomrule
\end{tabular}
\label{Table1}
\end{table*}


EFL also incorporates the sparsification and quantization to compress both the upstream (from clients to the server) and the downstream (from the server to clients) communications. It is not economical to only communicate the fraction of largest elements at full precision as regular top-$k$ sparsification \cite{aji2017sparse}. As a result, EFL quantizes the remaining top-$k$ elements of the sparsified updates to the mean population magnitude, leaving the updates with a ternary tensor containing $\{-\mu, 0, \mu\}$, which is summarized in Algorithm \ref{AL2}.

\subsection{Convergence Analysis}
Five assumptions are made to help analyze the convergence behaviors of EFL algorithm.

\begin{assumption}
\label{AS1}
(L-smoothness) $\widetilde{F}_1,...,\widetilde{F}_N$ are L-smooth, and F is also L-smooth.
\end{assumption}
\begin{assumption}
\label{AS2}
(Strong convexity) $\widetilde{F}_1,...,\widetilde{F}_N$ are $\mu$-strongly convex, and F is also $\mu$-strongly convex.
\end{assumption}
\begin{assumption}
\label{AS3}
(Bounded variance) The variance of the stochastic gradients is bounded by $\mathbb{E}_{\xi}||g_t^k-\overline{g}_t^k||^2\leq \sigma_k^2$
\end{assumption}
\begin{assumption}
\label{AS4}
(Bounded gradient) The expected squared norm of the stochastic gradients at each client is bounded by $\mathbb{E}_{\xi}||g_t^k||^2\leq G^2$.
\end{assumption}
\begin{assumption}
\label{AS5}
(Bounded aggregation coefficient) The aggregation coefficient has an upper bound, which is given by $p_{\tau}^k\leq \theta p^k$.
\end{assumption}
Assuming $\mathbb{E}[p_{\tau}^k], \mathbb{E}[p_{\tau}^ks_{\tau}^k], \mathbb{E}[(p_{\tau}^k)^2s_{\tau}^k]$, and $\mathbb{E}[\sum_{k=1}^Np_{\tau}^k-2+\sum_{k=1}^Np_{\tau}^ks_{\tau}^k]$ exist for all rounds $\tau$ and clients $k$, and $\mathbb{E}[\sum_{k=1}^Np_{\tau}^ks_{\tau}^k]\neq 0$. The convergence bound can be derived as
\begin{theorem}
\label{TH1}
Under Assumptions \ref{AS1} to \ref{AS5}, for learning rate $\eta_{\tau}=\frac{16E}{\mu ((\tau+1)E+\gamma)\mathbb{E}[\sum_{k=1}^Np_{\tau}^ks_{\tau}^k]}$, the EFL satisfies
\begin{equation}
\label{E7}
    \mathbb{E}||\omega_{\tau E}^G-\omega^*||^2\leq \frac{C_{\tau}}{(\tau E+\gamma)^2}+\frac{H_{\tau}J}{\tau E+\gamma},
\end{equation}
where $\gamma=\max\{\frac{4E^2\theta}{\min_{\tau}\mathbb{E}[\sum_{k=1}^Np_{\tau}^ks_{\tau}^k]}, \frac{32E(1+\theta)L}{\mu \min_{\tau}\mathbb{E}[\sum_{k=1}^Np_{\tau}^ks_{\tau}^k]}\}$, $H_{\tau}=\sum_{t=0}^{\tau-1}\mathbb{E}[r_t]$, $r_t \in \{0,1\}$ indicates the ratio $\frac{\mathbb{E}[p_{\tau}^ks_{\tau}^k]}{p^k}$ has the same value for all $k$, $C_{\tau}=\max \{\gamma^2\mathbb{E}||\omega_0^G-\omega^*||^2,(\frac{16E}{\mu})^2\sum_{t=0}^{\tau-1}\frac{\mathbb{E}[B_t]}{(\mathbb{E}[\sum_{k=1}^Np_t^ks_t^k])^2}\}$, $B_t=\sum_{k=1}^N(p_t^k)^2s_t^k\sigma_k^2+2(2+\theta)L\sum_{k=1}^Np_t^ks_t^k D_k+(2+\frac{\mu}{2(1+\theta)L})E(E-1)G^2(\sum_{k=1}^Np_t^ks_t^k+\theta(\sum_{k=1}^Np_t^k-2)+\sum_{k=1}^Np_t^ks_t^K)+2EG^2\sum_{k=1}^N\frac{(p_t^k)^2}{p^k}s_t^k$,  $J=\max_{\tau}\{\frac{32E\sum_{k=1}^N\mathbb{E}[p_{\tau}^ks_{\tau}^k]}{D_k/\mu \mathbb{E}[\sum_{k=1}^Np_{\tau}^ks_{\tau}^k]}\}$.
\end{theorem}
Based on Theorem \ref{TH1}, $C_{\tau}=O(\tau)$, which means $\omega_{\tau E}^G$ will finally converge to a global optimal as $\tau\rightarrow \infty$ if $H_{\tau}$ increases sub-linearly with $\tau$. Table \ref{Table1} summaries the required number of communication rounds of SCAFFOLD\cite{karimireddy2020scaffold}, MIME \cite{karimireddy2020mime}, VRL-SGD\cite{liang2019variance}, FedAMP\cite{huang2021personalized} and EFL. The proposed EFL algorithm achieves tighter bound comparing with methods which assume $\mu$-strongly convex.
The proof of Theorem \ref{TH1} is summarized in the supplementary material.

\section{Impacts of Irregular Clients}
In this section, we investigate the impacts of clients' different behaviors including being inactive, incomplete, new client arrival and client departure.

\begin{figure*}[tb]
\centering 
\includegraphics[scale=0.34]{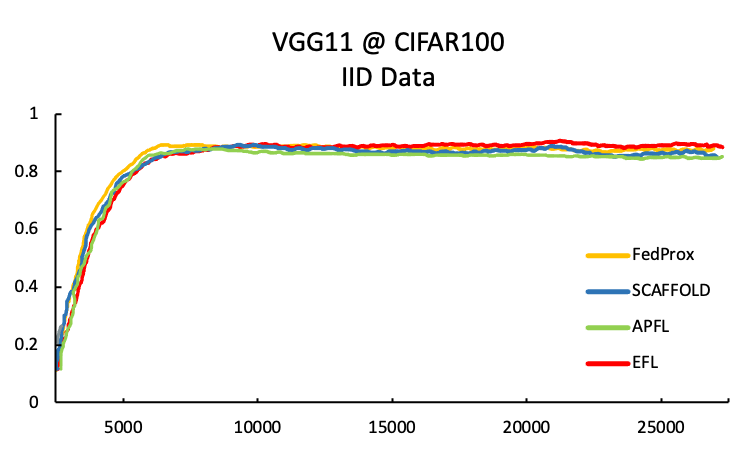}
\includegraphics[scale=0.34]{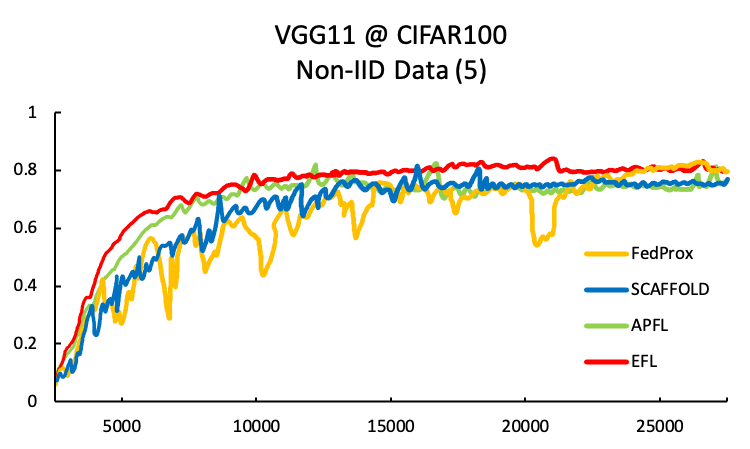}
\includegraphics[scale=0.34]{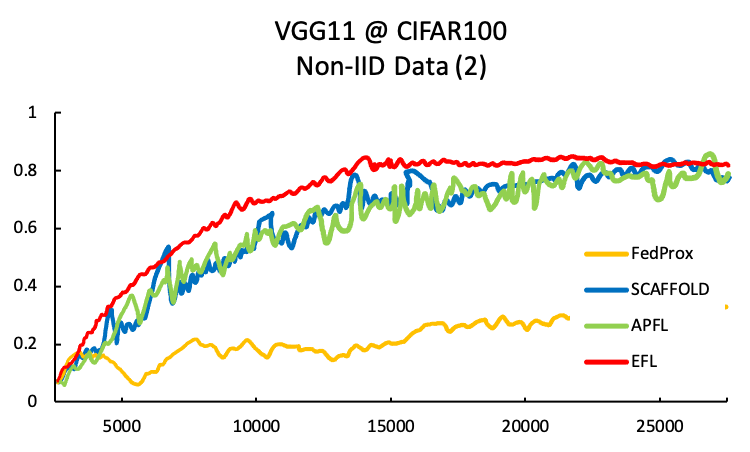}
\includegraphics[scale=0.34]{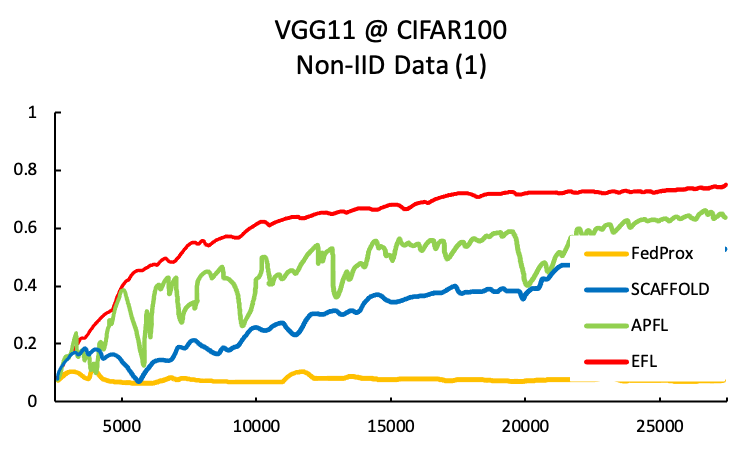}
\includegraphics[scale=0.34]{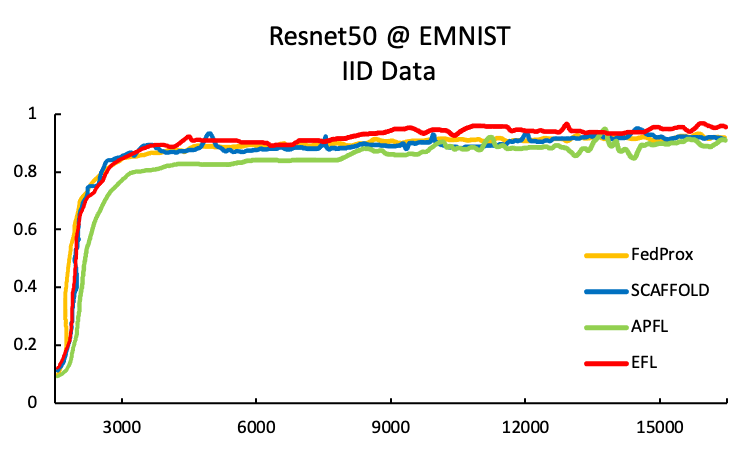}
\includegraphics[scale=0.34]{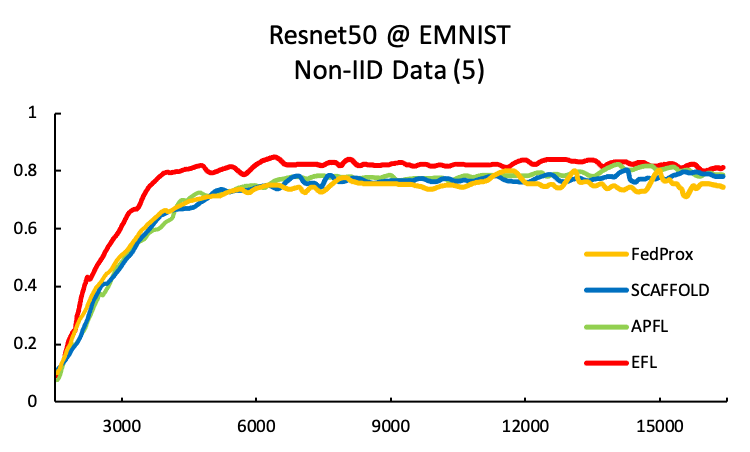}
\includegraphics[scale=0.34]{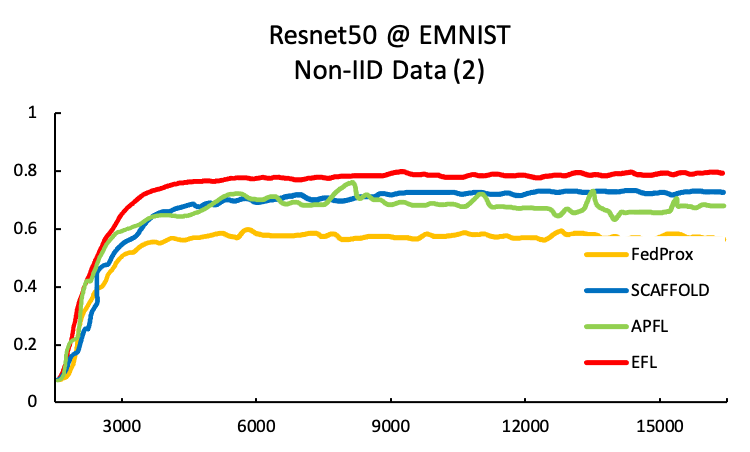}
\includegraphics[scale=0.34]{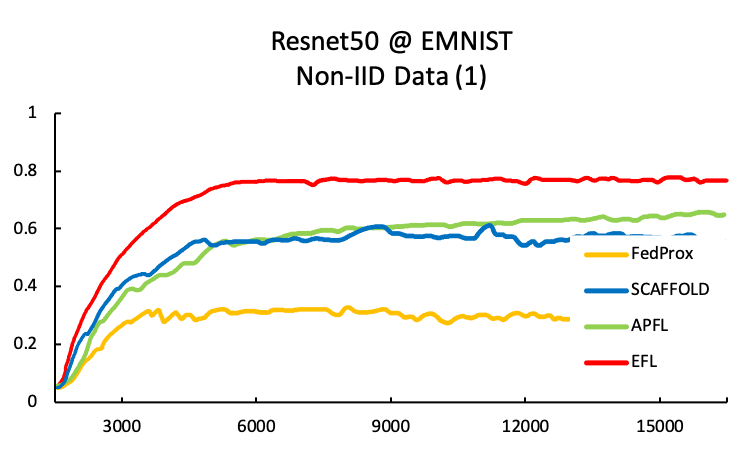}
\caption{Testing Accuracy-Communication Rounds comparisons of VGG11 on CIFAR100 and Resnet50 on EMNIST in a distributed setting for IID and non-IID data. In the non-IID cases, every client only holds examples from exactly $m$ classes in the dataset. All methods suffer from degraded convergence speed in the non-IID situation, but EFL is affected by far the least.}
\label{Figure1}
\end{figure*}

\subsection{Inactive Client}
If there exist inactive clients, the convergence rate changes to $O(\frac{\sum_{t=0}^{\tau-1}y_tJ}{\tau E+E^2})$. $y_t$ indicates if there are inactive clients in the $t$-th round. Furthermore, the term converges to zero if $y_t\in O(\tau)$, which means that a mild degree of inactive client will not discourage the convergence. A client can frequently become inactive due to the limited resources in reality. Permanently removing the client in this case may improve the model performance. Specially, we will remove the client if the system without this client leads to a smaller training loss when it terminates at the deadline $T$.

Suppose a client $a$ is inactive with the probability $0<y^a<1$ in each round, and let $f_0(\tau)$ be the convergence bound if we keep the client, $f_1(\tau)$ be the bound if it is abandoned at $\tau_0$. For $f_0$ with sufficiently many steps, the first term in Equation (\ref{E7}) shrinks to zero, and the second term converges to $y^aJ$. Thus, $f_0\approx y^aJ$, and we can obtain that $f_1{\tau}=\frac{\widetilde{C}_{\tau}}{(E(\tau-\tau_0)+\widetilde{\gamma})^2}$ for some $\widetilde{C}_{\tau}$ and $\widetilde{\gamma}$, thus we have
\begin{corollary} 
\label{co1}
An inactive client a should be abandoned if
\begin{equation}
\label{E8}
    y^aJ>f_1(T).
\end{equation}
Assuming $C_{\tau}\approx \widetilde{C}_{\tau}=\tau C$ and $\gamma\approx\widetilde{\gamma}$, i.e., the removed client does not significantly affect the overall SGD variance and the degree of non-IID, then Equation (\ref{E8}) can be formulated as
\begin{equation}
    y^k>O(\frac{\frac{C}{J}}{TE}).
\end{equation}
\end{corollary}
From Corollary \ref{co1}, the more epochs the training on the local client, the more sensitive it is to the in-activeness.

\subsection{Incomplete Client}
Based on Theorem \ref{TH1}, the convergence bound is controlled by the expectation of $p_{\tau}^k$ and its functions. EFL allows clients to upload partial work with adaptive weight $p_{\tau}^k=\frac{Ep^k}{s_{\tau}^k}$. It assigns a greater aggregation coefficient to clients that complete fewer local epochs, and turns out to guarantee the convergence in the non-IID setting. The resulting convergence bound follows $O(\frac{E^5\sum_{t=0}^{\tau-1}\sum_k^Np^k\mathbb{E}[1/s_t^k]+E^2\sum_{t=0}^{\tau-1}\sum_k^N(p^k\sigma_k)^2\mathbb{E}[1/s_t^k]}{(\tau E+E^2)^2})$.

The reason for enlarging the aggregation coefficient lies in Equation (\ref{E6}) that increasing $p_{\tau}^k$ is equivalent to increasing the learning rate of client $k$. By assigning clients that complete fewer epochs a greater aggregation coefficient, these clients effectively run further in each local step, compensating for less epochs.

\begin{table}[tb]
\centering
\caption{BMTA for the non-IID data setting}
\begin{tabular}{ccccc} 
\toprule
Methods & MNIST & CIFAR100 & Sent140 & Shakes. \\
    \midrule
    FedAvg & $98.30$ & $2.27$ & $59.14$ & $51.35$\\
    \cline{1-5}
    FedGATE  & $\textbf{99.15}$ & $80.94$ & $68.84$ &$54.71$ \\
    VRL-SGD & $98.86$ & $2.81$ & $68.62$ & $52.33$ \\
    \cline{1-5}  
    APFL & $98.49$ & $77.19$ & $68.81$ &$55.27$ \\
    FedAMP & $99.06$ & $81.17$ & $\textbf{69.01}$ & $58.42$ \\
    \cline{1-5}
    EFL& $99.10$ & $\textbf{81.38}$ & $68.95$ &$\textbf{60.49}$ \\
\bottomrule
\end{tabular}
\label{Table3}
\end{table}

\subsection{Client Departure}
If $k$-th client quits at $\tau_0<T$, no more updates will be received from it, and $s_{\tau}^k=0$ for all $\tau>\tau_0$. As a result, the value of ratio $\frac{\mathbb{E}[p_{\tau}^ks_{\tau}^k]}{p^k}$ is different for different $k$, and $r_{\tau}=1$ for all $\tau>\tau_0$. According to Theorem \ref{TH1}, $\omega_{\tau E}^G$ cannot converge to the global optimal $\omega^*$ as $H_{T}\geq T-\tau_0$. Intuitively, a client should contribute sufficiently many updates in order for its features to be captured by the trained model in the non-IID setting. After a client leaves, the remaining training steps will not keep much memory as it runs more rounds. Thus, the model may not be applicable to the leaving client, especially when it leaves early in the training ($\tau_0\ll T$), which indicates that we may discard a departing client if we cannot guarantee the trained model performs well on it, and the earlier a client leaves, the more likely it should be discarded. 

However, removing the departing client (the $a$ client) from the training may push the original learning objective $F=\sum_{k=1}^N p^k\widetilde{F}_k$ towards the new one $\hat{F}=\sum_{k=1,k\neq a}^N p^k\widetilde{F}_k$, and the optimal weight $\omega^*$ will also shift to some $\hat{\omega}^*$ that minimizes $\hat{F}$. There exists a gap between these two optimal, which further adds an additional term to the convergence bound obtained in Theorem \ref{TH1}, and a sufficient number of updates are required for $\omega_{\tau E}^G$ to converge to the new optimal $\hat{\omega}^*$.

\subsection{Client Arrival}
The same argument holds when a new client joins in the training, which requires changing the original global objective to include the loss on the new client's data. The learning rate also needs to be increased when the objective changes. Intuitively, if the shift happens at a large time $\tau_0$, where $\omega_{\tau E}^G$ approaches to the old optimal $\omega^*$ and $\eta_{\tau_0}$ is close to zero, reducing the latest differences $||\omega_{\tau E}^G-\hat{\omega}^*||^2\approx||\omega^*-\hat{\omega}^*||^2$ with a small learning rate is inapplicable. Thus, a greater learning rate should be adopted, which is equivalent to initiate a fresh start after the shift, and there still needs more updating rounds to fully address the new client.

We also present the bounds of the additional term due to the objective shift as
\begin{theorem}
\label{TH2}
For the global objective shift $F\rightarrow\hat{F}$, $\omega^*\rightarrow\hat{\omega}^*$, let $\hat{D}_k=F_k(\hat{\omega}^*)-F_k^*$ quantify the degree of non-IID data with respect to the new objective. If the client $a$ quits the system,
\begin{equation}
    \label{E10}
    ||\omega^*-\hat{\omega}^*||^2\leq \frac{8Ln_a^2\hat{D}_a}{\mu^2n^2}.
\end{equation}
If the client $a$ joins the system,
\begin{equation}
    \label{E11}
    ||\omega^*-\hat{\omega}^*||^2\leq \frac{8Ln_a^2\hat{D}_a}{\mu^2(n+n_a)^2},
\end{equation}
where $n$ is the total number of samples before the shift.
\end{theorem}
It can be concluded that the bound reduces when the data becomes more IID, and when the changed client owns fewer data samples.


\section{Experiments}

In this section, we first demonstrate the effectiveness and efficiency of EFL in the non-IID data setting, and compare with several baseline algorithms. Then, we show the robustness of EFL on the low participation rate challenge.

\subsection{Experimental Settings}
\label{Section 4.1}
Both convex and non-convex models are evaluated on a number of benchmark datasets of federated learning. Specifically, we adopt MNIST \cite{lecun1998gradient}, EMNIST \cite{cohen2017emnist} dataset with Resnet50 \cite{he2016deep}, CIFAR100 dataset \cite{krizhevsky2009learning} with VGG11 \cite{simonyan2014very} network, Shakespeare dataset with an LSTM \cite{mcmahan2017communication}to predict the next character, Sentiment140 dataset \cite{go2009twitter} with an LSTM to classify sentiment and synthetic dataset with a linear regression classifier.

Our experiments are conducted on the \textit{TensorFlow} platform running in a Linux server. For reference, statistics of datasets, implementation details and the anonymized code are summarized in supplementary material.


\begin{figure*}[tb]
\centering 
\includegraphics[scale=0.34]{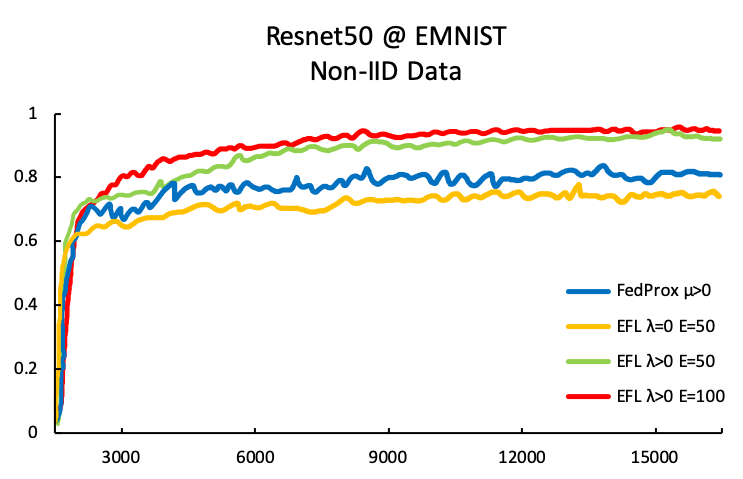}
\includegraphics[scale=0.34]{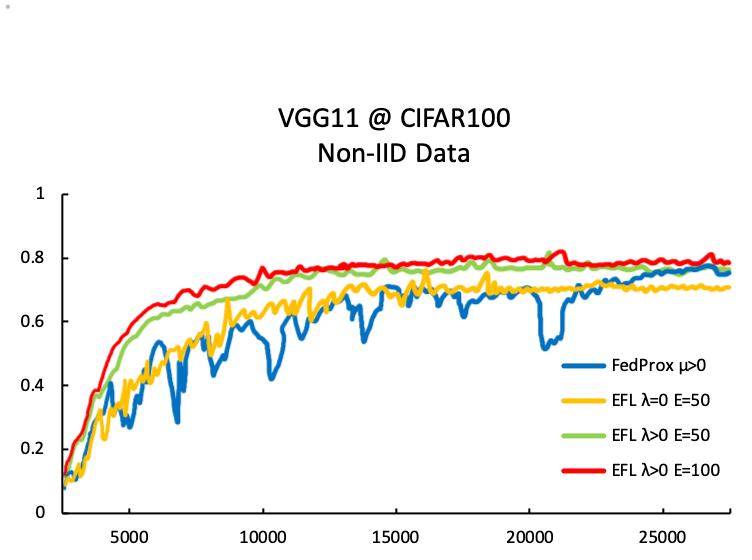}
\includegraphics[scale=0.34]{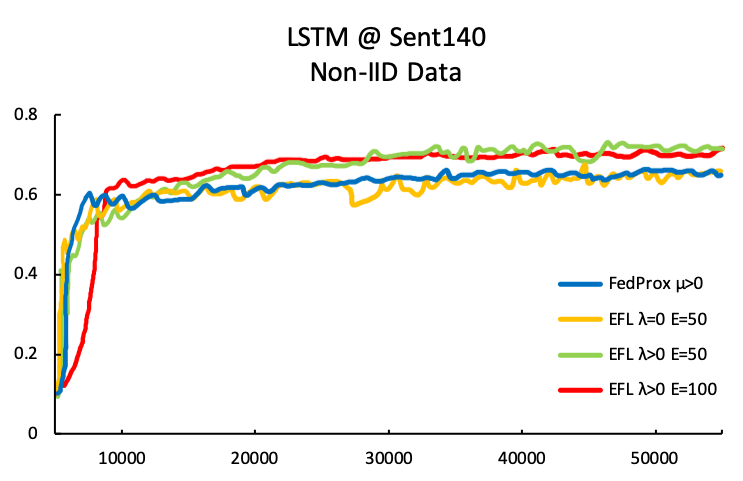}
\includegraphics[scale=0.34]{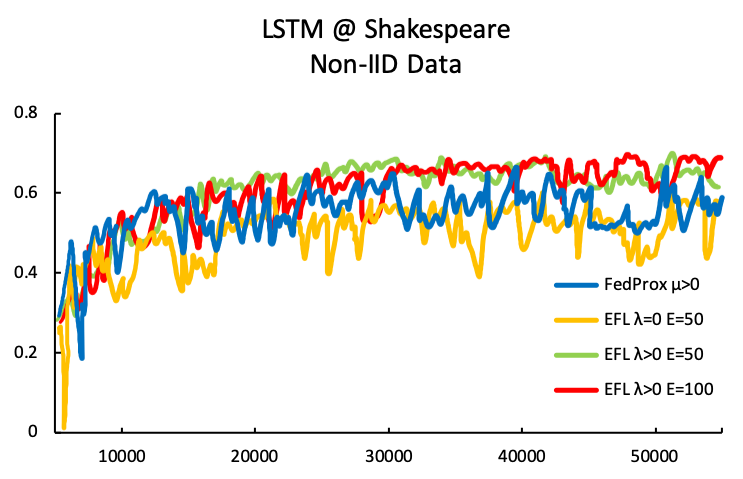}
\includegraphics[scale=0.34]{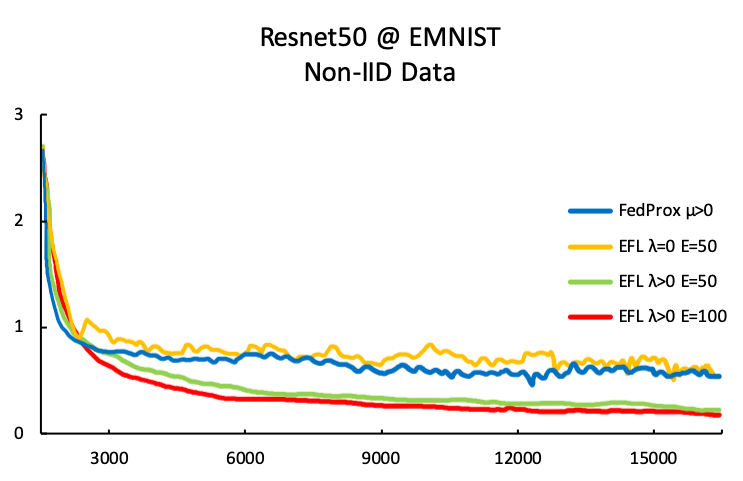}
\includegraphics[scale=0.34]{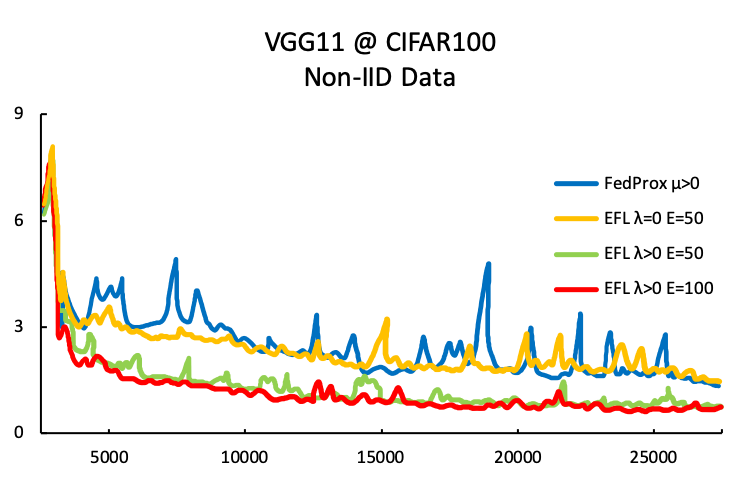}
\includegraphics[scale=0.34]{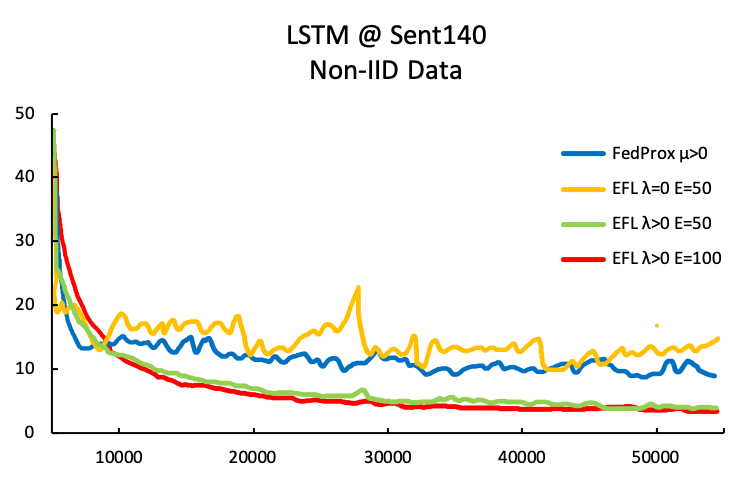}
\includegraphics[scale=0.34]{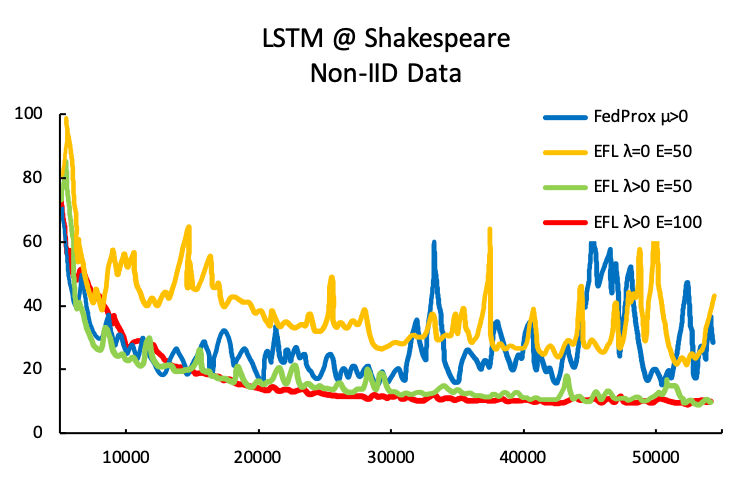}
\caption{The first row shows Testing Accuracy-Communication Rounds comparison and the second row shows Training Loss-Communication Rounds comparison in non-IID setting. EFL with elastic term stabilizes and improves the convergence of the algorithm.}
\label{Figure2}
\end{figure*}

\subsection{Effects of Non-IID Data}

We run experiments with a simplified version of the well-studied 11-layer VGG11 network, which we train on the CIFAR100 dataset in a federated learning setup using 100 clients. For the IID setting, we split the training data randomly into equally sized shards and assign one shard to every clients. For the non-IID ($m$) setting, we assign every client samples from exactly $m$ classes of the dataset. 
We also perform experiments with Resnet50, where we train on EMNIST dataset under the same setup of the federated learning environment. Both models are trained using SGD.

Figure \ref{Figure1} shows the convergence comparison in terms of gradient evaluations for the two models using different algorithms. FedProx \cite{li2018federated} incorporates a proximal term in local objective to improve the model performance on the non-IID data, SCAFFOLD \cite{karimireddy2020scaffold} adopts control variate to alleviate the effects of data heterogeneity, and APFL \cite{deng2020adaptive} learns personalized local models to mitigate heterogeneous data on clients.

\begin{figure}[tb]
\centering 
\includegraphics[scale=0.33]{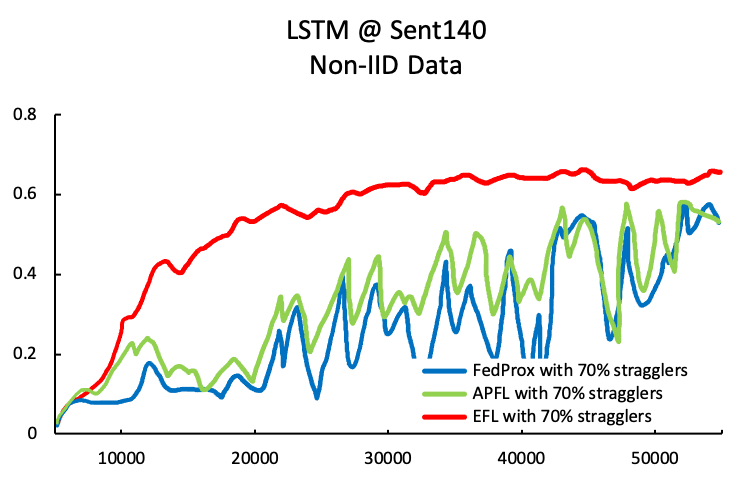}
\includegraphics[scale=0.33]{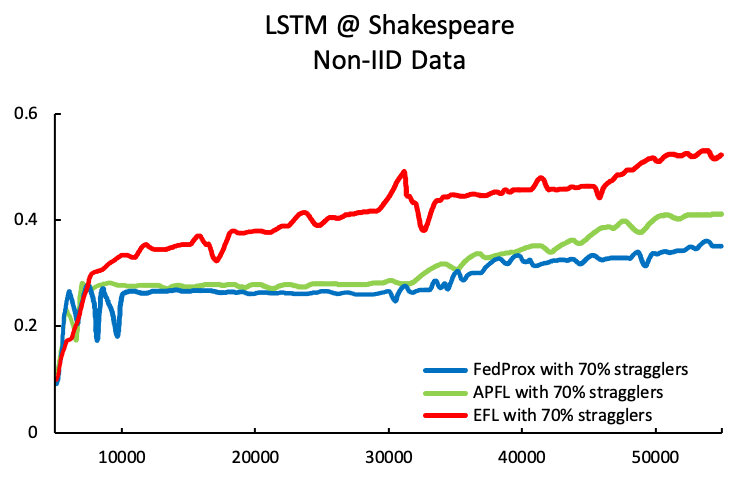}
\caption{Testing Accuracy-Communication Rounds comparisons among different algorithms in low participation rate. EFL utilizes incomplete updates from stragglers and is robust to the low participation rate.}
\label{Figure3}
\end{figure}

We observe that while all methods achieve comparably fast convergence in terms of gradient evaluations on IID data, they suffer considerably in the non-IID setting. From left to right, as data becomes more non-IID, convergence becomes worse for FedProx, and it can diverge in some cases. SCAFFOLD and APFL exhibit its ability in alleviating the data heterogeneity, but are not stable during training. As this trend can be observed also for Resnet50 on EMNIST case, it can be concluded that the performance loss that is originated from the non-IID data is not unique to some functions.

Aiming at better illustrating the effectiveness of the proposed algorithm, we further evaluate and compare EFL with the state-of-the-art algorithms, including FedGATE \cite{haddadpour2021federated}, VRL-SGD \cite{liang2019variance}, APFL \cite{deng2020adaptive} and FedAMP \cite{huang2021personalized} on MNIST, CIFAR100, Sentiment140, and Shakespeare dataset. The performance of all the methods is evaluated by the best mean testing accuracy (BMTA) in percentage, where the mean testing accuracy is the average of the testing accuracy on all participants. For each of datasets, we apply a non-IID data setting.

Table \ref{Table3} shows the BMTA of all the methods under non-IID data setting, which is not easy for vanilla algorithm FedAvg. On the challenging CIFAR100 dataset, VRL-SGD is unstable and performs catastrophically because the models are destroyed such that the customized gradient updates in the method can not tune it up. APFL and FedAMP train personalized models to alleviate the non-IID data, however, the performance of APFL is still damaged by unstable training. FedGATE, FedAMP and EFL achieve comparably good performance on all datasets.

\subsection{Effects of the Elastic Term}

EFL utilizes the incomplete local updates, which indicates that clients may perform different amount of local work $s_{\tau}^k$, and this parameter together with the elastic term scaled by $\lambda$ affect the performance of the algorithm. However, $s_{\tau}^k$ is determined by its constrains, i.e., it is a client specific parameter, EFL can only set the maximum number of local epochs to prevent local models drifting too far away from the global model, and tune a best $\lambda$. Intuitively, a proper $\lambda$ restricts the optimization trajectory by limiting the most informative parameters' change, and guarantees the convergence.

We explore impacts of the elastic term by setting different values of $\lambda$, and investigate whether the maximum number of local epochs influences the convergence behavior of the algorithm. Figure \ref{Figure2} shows the performance comparison on different datasets using different models. We compare the result between EFL with $\lambda=0$ and EFL with best $\lambda$. For all datasets, it can be observed that the appropriate $\lambda$ can increase the stability for unstable methods and can force divergent methods to converge, and it also increases the accuracy in most cases. As a result, setting $\lambda\geq0$ is particularly useful in the non-IID setting, which indicates that the EFL benefits practical federated settings.




\subsection{Robustness of EFL}
\label{Section 4.3}
Finally, in Figure \ref{Figure3}, we demonstrate that EFL is robust to the low participation rate. In particular, we track the convergence speed of LSTM trained on Sentiment140 and Shakespeare dataset. It can be observed that reducing the participation rate has negative effects on all methods. The causes for these negative effects, however, are different: In FedAvg, the actual participation rate is determined by the number of clients that finish the complete training process, because it does not include the incomplete updates. This can steer the optimization process away from the minimum and might even cause catastrophic forgetting. On the other hand, low participation rate reduces the convergence speed of EFL by causing the clients residuals to go out sync and increasing the gradient staleness. The more rounds a client has to wait before it is selected to participate in training again, the more outdated the accumulated gradients become.

\section{Conclusion}

In this paper, we propose EFL as an unbiased FL algorithm that can adapt to the statistical diversity issue by making the most informative parameters less volatile. EFL can be understood as an alternative paradigm for fair FL, which tackles bias that is originated from the non-IID data and the low participation rate. Theoretically, we provide convergence guarantees for EFL when training on the non-IID data at the low participation rate. Empirically, experiments support the competitive performance of the algorithm on the robustness and efficiency.

\bibliographystyle{named}
\bibliography{mybibliography}

\end{document}